\def\year{2020}\relax
\documentclass[letterpaper]{article} 
\usepackage{aaai20}  
\usepackage{times}  
\usepackage{helvet} 
\usepackage{courier}  
\usepackage[hyphens]{url}  
\usepackage{graphicx} 
\usepackage{graphicx}  
\usepackage{natbib}

\nocopyright

\usepackage{times}

\usepackage{latexsym}
\usepackage{amsmath}
\usepackage{url}
\usepackage{comment}
\usepackage{paralist}
\usepackage{soul}

\makeatletter
\newcommand{\thickhline}{%
	\noalign {\ifnum 0=`}\fi \hrule height 1pt
	\futurelet \reserved@a \@xhline
}

\title{Controllable Sentence Simplification: \\ Employing Syntactic and Lexical Constraints }

\author{Jonathan Mallinson {\normalfont and} Mirella Lapata \\
Institute for Language, Cognition and Computation \\
School of Informatics, University of Edinburgh \\
10 Crichton Street, Edinburgh EH8 9AB \\
j.mallinson@ed.ac.uk  mlap@inf.ed.ac.uk}

\begin{document}

\maketitle
\bibliographystyle{aaaime} 
\begin{abstract}
Sentence simplification aims to make sentences easier to
read and understand.  Recent approaches have shown promising
results with sequence-to-sequence models which have been
developed assuming homogeneous target audiences.  In this
paper we argue that different users have different
simplification needs (e.g.,~dyslexics vs. non-native
speakers), and propose CROSS, a
\textbf{C}ont\textbf{RO}llable \textbf{S}entence
\textbf{S}implification model, which allows to control both
the \emph{level} of simplicity and the \emph{type} of the
simplification. We achieve this by enriching a
Transformer-based architecture with syntactic and lexical
constraints (which can be set or learned from data).
Empirical results on two benchmark datasets show that
constraints are key to successful simplification, offering
flexible generation output.
\end{abstract}
\section{Introduction}
Sentence simplification aims to reduce the linguistic complexity of a
text whilst retaining most of its meaning. It has been the subject of
several modeling efforts in recent years due to its relevance to
various applications
\cite{siddharthan2014survey,shardlow2014survey}. Examples include the
development of reading aids for individuals with autism
\cite{evans2014evaluation}, aphasia \cite{carroll1999simplifying},
dyslexia \cite{rello2013simplify}, and population groups with
low-literacy skills
\cite{watanabe2009facilita},
such as children and non-native speakers.

Modern approaches \cite{zhang2017sentence,vu2018sentence,guo2018dynamic,zhao2018integrating}
view the simplification task as monolingual text-to-text rewriting and
employ the very successful encoder-decoder neural architecture
\cite{bahdanau2015neural,sutskever2014sequence}. In contrast to
traditional methods which target individual aspects of the
simplification task such as sentence splitting
\cite{carroll1999simplifying,chandrasekar1996motivations},
inter alia) or the substitution of complex words with simpler ones
\cite{Devlin:1999,kaji-EtAl:2002:ACL}, neural models have no
special-purpose mechanisms for ensuring how to best simplify
text. They rely on representation learning to \emph{implicitly} learn
simplification rewrites from data, i.e.,~examples of complex-simple
sentence pairs.

In this paper, we propose a user-centric simplification model
which draws on the advantages of the encoder-decoder
architecture but can also \emph{explicitly} model rewrite
operations, such as lexical and syntactic simplifications, and
as a result generate output according to
specifications. Although many simplification systems
\cite{zhu2010monolingual,kauchak2013improving,zhang2017sentence}
are intended as general purpose, different target populations
may have different needs \cite{siddharthan2014survey}. For
instance, whether or not the syntax should be simplified
depends on the reader: those affected by aphasia benefit from
simpler syntax, while dyslexics have trouble processing long
and infrequent words
\cite{rello2013frequent,shewan1971effects}. It is therefore
beneficial to have a model which can be easily adapted for
particular users or user populations without being redesigned
every time from scratch.

Our simplification model adopts the Transformer architecture
\cite{vaswani2017attention} which has become state-of-the-art in
machine translation \cite{bojar-EtAl:2018:WMT1} and relies entirely on
self-attention to compute representations of its input and output
without using recurrent or convolutional neural networks. Our
innovation is to enrich a Transformer-based sequence-to-sequence model
with syntactic and lexical constraints which allow the user to control
both the \emph{level} of simplicity and the \emph{type} of
simplification. Importantly, this requires no additional annotation
(e.g., for grade levels) and the constrains are applied post training,
allowing one model to be used across datasets and tasks.  We enable
the model to make decisions about which words or syntactic structures
to replace by enriching the training data with explicit information
pertaining to lexical substitution and syntactic simplification. For
example, we can mark words as to \emph{keep} or \emph{substitute}, or
append a high-level level syntactic description (a template) to the
source and target sentence. At test time, the user provides their
constraints and the decoder must first decode the syntax of the target
sentence before decoding the lexical tokens.

We evaluate our system on two publicly available datasets collected
automatically from Wikipedia
\cite{woodsend2011learning,kauchak2013improving,zhu2010monolingual}
and human-authored news articles \cite{xu2015problems} and report
results using automatic and human evaluation.  By comparing our
constrained model against non-constrained variants we show that
constraints are key to successful simplification, offering generation
flexibility and controllable output.  Our contributions in this paper
are three-fold: (1)~we show that adding lexical and syntactic
constraints to a Transformer produces state-of-the-art simplification
results; (2)~these constraints allow users to adapt the model to their
personal needs; and (3)~we conduct a comprehensive evaluation and
comparison study which highlights the merits and shortcomings of
various recently proposed simplification models on two datasets.

\section{Related work}

Our model resonates the recent trend of developing simplification
models using neural architectures based on the encoder-decoder
paradigm. It also agrees with previous work in acknowledging that both
lexical and syntactic information is important in creating simplified
text.

One of the first neural network approaches to simplification
was presented by \citet{zhang2017sentence} who use an
encoder-decoder LSTM, trained using reinforcement learning, to
optimize for grammaticality, simplicity, and adequacy. They
also propose an extension which ensembles their basic model
with a lexical simplification
component. \citet{vu2018sentence} augment an encoder-decoder
model with the Neural Semantic Encoder
\cite{munkhdalai2017neural}: a variable sized memory that
updates over time through the use of read, compose, and write
operations. This increased capacity allows for better tracking
of long range dependencies encountered within sentence
simplification.

\citet{guo2018dynamic} use multi-task learning to augment the
limited amount of simplification training data.  In addition
to training on complex-simple sentence pairs, their model
employs paraphrases, created automatically using machine
translation, and entailment
pairs. \citet{zhao2018integrating} are
closest to our work; they augment a Transformer-based
simplification model with lexical rules obtained from simple
PPDB \cite{ganitkevitch2013ppdb}, a subset of PPDB which has
been automatically annotated with a simplicity score. A memory
unit is added to the model which holds the applicable PPDB
rules and a new loss rewards the model using rules from simple
PPDB. Although the backbone of our model is also a
Transformer, our aim is to develop a simplification system
capable of adapting to the individual needs of specific users.

In recent years there has been increased interest in controlling the
output of sequence-to-sequence models. Previous work has focused on
controlling the length and content of summaries
\citep{kikuchi2016controlling,fan2017controllable}, politeness in
machine translation \cite{Sennrich:ea:2016}, and style
\cite{ficler2017controlling}.  \citet{scarton-specia-2018-learning}
develop a text simplification model that controls the grade level of
the output.  They train a text sequence-to-sequence model on
Newsela, attaching tags which specify the output sentences' grade
level. \citet{nishihara2019controllable} expand upon this work by
weighting the loss function to favor the generation of certain
words. In this way, they can train different models with different
output lexical preferences.  As both approaches require explicit
grade level annotations, they cannot be used with Wikipedia based
simplification datasets.

  In contrast, our work requires no grade level annotation, and user
  control is applied at test time, allowing us to train only one
  model. Our work draws inspiration from
\citet{grangier2017quickedit} who post-edit the output of machine
translation under the assumption that a human modifies a sentence by
marking tokens they would like the system to change. Our model also
controls simplification by taking as input both the sentence and
change markers for it.  However, we allow for a wider spectrum of
rewrite operations than \citet{grangier2017quickedit} who focus solely
on deletion and do not take syntax into account. \citet{C18-1021}
notably acknowledge the fact that there is no one-size-fits-all
solution to text simplification and develop a tool which can be
personalized to user needs and adapt over time. Their system decides
whether a word (in context) poses difficulty to the reader and
suggests lexical substitutions.

Earlier approaches to simplification rely heavily on syntax,
either by developing rule-based components
\cite{chandrasekar1996motivations} or models which operate
over parse trees and learn a mapping from complex to simpler
structures
\cite{xu2016optimizing,woodsend2011learning}. Our model
is informed by syntax, however, only indirectly since
generation proceeds sequentially
token-by-token. \citet{iyyer2018adversarial} learn how to
generate paraphrases subject to a syntactic template.  We
adapt their template extraction method to the simplification
task and incorporate it in our model.

\section{Model Description}
\label{sec:model-description}

In this paper, we devise a model that adapts to the user's
simplification needs. The main idea is to equip a neural
encoder-decoder model with constraints. 
The model still learns how to simplify from data, i.e.,~pairs
of source (complex) and target (simple) sentences which are
additionally annotated with change markers (e.g.,~indicating
which words to replace, which syntactic constructs to delete)
and takes these into account while generating simplifications.

\subsection{Transformer} 

We will first define a basic encoder-decoder model for
sentence simplification and then explain how to add
constraints. Given a complex sentence
$X = (x_1, x_2, . . . , x_{|X|} )$, our model learns to
predict its simplified target
$Y = (y_1, y_2, . . . , y_{|Y|})$. Inferring the target $Y$
given source $X$ can be modeled as a sequence-to-sequence
learning problem. We adopt Transformer's multi-layer and
multi-head attention architecture \cite{vaswani2017attention}.

The hidden state~$h_l^i$, at time step~$i$ in layer~$l$ in the
Transformer encoder is calculated from all hidden states of
the previous layer, as seen below:
\begin{equation}
h^l_i = h^{l-1}_i + f(\text{self-attention}( h^{l-1}_i))
\label{eq.t.l}
\end{equation}
where~$f$ is a feed-forward network using ReLU and layer normalization \cite{ba2016layer}. In the input
layer, $h^0_i$ is calculated as:
\begin{equation}
h^0_i = E_{x_i} + e_{pos,i}
\end{equation}
where~$E$ is the word embedding matrix and $e_{pos,i}$ are positional
embeddings.

Analogously, the decoder also consists of multiple layers,
which apply self-attention. However, the decoder has an
additional attention network, inserted after the
self-attention network, which attends over the source sentence
hidden states. Each hidden state~$h_l^i$ in the final
layer~$L$, is fed through a softmax ranging over the target
word vocabulary.

\begin{table*}[t]
	\small\centering
	{
		\begin{tabular}{l@{\hspace{0.8ex}}l@{\hspace{1.4ex}}} \thickhline 
		
			Tokens   & take the square root of the variance . \\\thickhline
			Linearized  & ROOT( take OBJ( DET( the ) AMOD( square ) NMOD( variance CASE( of ) DET( the ) ) ) PUNCT( . ) ) \\
			Template  &  OBJ( AMOD( d0 ) DET( d0 ) NMOD( d1) ) PUNCT( ) \\  
			
			Input/Output & OBJ( AMOD( d0 ) DET( d0 ) NMOD( d1) OBJ) PUNCT( )  $\vert \vert \vert$  take the square root of the variance . \\ 
			Constraints &  ROOT(OBJ, PUNCT), OBJ(AMOD, DET, NMOD), PUNCT() \\ \thickhline 
	\end{tabular}}
	\caption{Example of source sentence with  \emph{linearized}
		parse, \emph{template}, constraints extracted from the
		template, and input provided to our model (for training). To convert from a linearized parse to a template,
		first the dependents are ordered, then the opening and closing
		brackets are matched together (excluded for brevity). Finally, we remove levels lower
		than 2 and instead replace them with the \emph{d*} token which
		represents the maximum depth of the child.} 
	\label{tbl.syntax}
\end{table*}

\subsection{Lexical Constraints} 
Lexical substitution, the replacement of complex words with simpler
alternatives, is an integral part of sentence simplification and has
been the subject of much previous work
\cite{specia2012semeval,E17-2006,C18-1019,Yatskar:ea:10,Devlin:1999,inui-EtAl:2003:PARAPHRASE,kaji-EtAl:2002:ACL}.
We enrich the encoder of the Transformer with lexical constraints, by
adding indicator features to each word embedding, specifying if the
token should be kept.  We employ three indicator types:
\begin{enumerate}
	\item The token should be replaced; during training this is set if the
	token does not appear in the target sentence;
	\item The token should be kept; during training this is set if the
	token  is in the target sentence;
	\item There is no preference for  the token to be kept or replaced;
	during training half of all tokens are randomly assigned this
	value. 
\end{enumerate}

Unlike \citet{grangier2017quickedit}, we do not require tokens
in the source and target to constitute an exact
match. Instead, we apply constraints more flexibly, and mark
tokens (to be replaced or kept) as long as their stems
match. Indicator features are added to the word embedding and
positional encoding, as seen in the equation below:
\begin{equation}
h^0_i = E_{x_i} + e_{pos,i} + cw_i 
\label{eq.l.s}
\end{equation}
where $cw_i$ are indicator features learnt during training. We
also restrict the generation of complex words; during decoding
we use constrained beam search, where complex words are given
zero probability \cite{post2018fast}.

At \emph{test time}, the user can control the model's output
simply by (1)~striking out tokens they wish to discard; (2)
marking tokens they want to keep; or~(3) leaving tokens
unmarked. These could be words that an aphasic reader has
trouble understanding, or a second language learner is not
familiar with.  For example in the sentence
``\textsl{Dextromethorphan \st{occurs} as a \textbf{white}
	\textbf{powder}}'', \textit{occurs} should be replaced and
\textsl{white powder} should be preserved.  Lists of complex
words can be provided to the model in two formats: as a
\emph{dictionary} of complex and corresponding simple words or
as a \emph{list} of complex words.  When a dictionary is
available, we mark for replacement all complex words and
during decoding we constrain the output to include only words
which appear as simplifications. When a list is provided, we
again mark for replacement complex words and leave it up to
the model to decide what to simplify.

In experiments we used the simplification
dictionary\footnote{\url{http://www.spencerwaterbed.com/soft/			simple/about.html}} provided by the Wikipedia editor
``SpencerK'' (Spencer Kelly). Due to the limited size of this
dictionary, we combine it with an automatically created
simplification dictionary, learnt from the training
data. Word alignments, produced using GIZA++ \cite{och03:asc},
were used to create phrase tables, which we treat as a
simplification dictionary (\textsl{abandon} $\rightarrow$
\textsl{leave}, \textsl{replenished} $\rightarrow$
\textsl{filled}, \textsl{fraudulent} $\rightarrow$
\textsl{fake}; see the supplementary
material for more examples).

In addition we use a fairly inexpensive approach to learn a list of
complex words from training data. We calculate the relative
probability that a word appears in the simple and complex
corpora:
\begin{equation}
\label{eq:ratio}
\mathrm{Complexity}(word) = \frac{P(word|complex)}{P(word|simple)}
\end{equation}
Using Equation~\eqref{eq:ratio}, we order all words in the
training set with~$\mathrm{Complexity}(word){>}1$ and take the
first~$N$ words to produce the complex list (e.g.,
\textsl{cavalier}, \textsl{offbeat}, \textsl{insofar}; see the
supplementary for more examples).

\subsection{Syntactic Constraints}

Syntactic simplification aims to reduce the syntactic complexity of a
text while preserving its meaning and information content. Although
the bulk of previous work has focused on sentence splitting, namely
rewriting a complex sentence into multiple simpler sentences
\cite{carroll1999simplifying,chandrasekar1996motivations},
other operations which reduce syntactic complexity involve rendering
passive voice into active, simplifying relative clauses and
coordination, as well reordering constituents or deleting them.

Syntax is introduced to our model by annotating the complex source and
simplified target with high level syntactic descriptions (aka
templates).  Templates are induced from the training corpus by parsing
source and target sentences with a universal dependencies parser 
\cite{straka2018udpipe}. An example of a parse can be seen in Table
\ref{tbl.syntax}. Dependency parses are further linearized and we
extract a template corresponding to the top two levels of the parse.
Templates are appended to the front of the source and target
sentences. Once the model is trained on this template-enriched corpus,
the decoder must first generate a target template and then decode the
string.

The annotation process described above renders the model
syntax-aware. Analogously to the lexical constraints, a globally
constraint variant of beam search is used at test time and syntactic
indicator features (i.e.,~replace, keep, don't care) are added to the
encoder.  To reduce sparsity, a Markovian assumption is applied to the
templates.  Each constraint consists of one parent and its children as
found within the template (see Table~\ref{tbl.syntax} for
examples). Unlike lexical constraints, which are applied at the token
level, syntactic constraints are applied at the rule level.  At test
time, the user provides a list of constraints the system must adhere
to. The list is used to mark the input syntax and to constrain the
decoder's output.  For example, applying the constraint Root(nsubj
nmod nmod advcl) $\rightarrow$ Root(nsubj nmod advcl) to the source
sentence ``\textsl{She remained in the United States until 1927 when
  she and her husband returned to France.}'' produces the
simplification ``\textsl{She remained in the USA until she returned to
  France with her husband in 1927.}''  As with lexical constraints, we
provide syntactic simplifications in two formats. As a list of
synchronous grammar rules (see Table~\ref{tbl.sp}) or a list of
complex rules which the output must avoid (see Table~\ref{tbl.su};
Newsela and WikiLarge are benchmark datasets we experimented with; see
 next section for details).

\begin{table}[t]
	\small \centering
	\begin{tabular}{l@{\hspace{1ex}}l@{\hspace{1ex}}} \thickhline 
		
		\multicolumn{1}{c}{Complex}   & \multicolumn{1}{c}{Simple}                              \\ \thickhline
		\hspace{-0.14cm}Root(conj,punct)                                               &  Root(punct)                  \\
		\hspace{-0.14cm}Root(obj,conj,punct)                                         &  Root(obj,punct)  \\
		\hspace{-0.14cm}Root(advcl,nmod,nmod,nsubj)                                &  Root(advcl,nmod,nsubj)   \\ \thickhline 
	\end{tabular}

	\caption{Example of synchronous grammar rules.}
	\label{tbl.sp}

\end{table}

\begin{table}[]
	\small
	\centering
	\begin{tabular}{l} \thickhline 
		
		\multicolumn{1}{c}{WikiLarge}                         \\ \thickhline
		Root(cop, det, nsubj, punct, vocative)               \\
		Root(avmod, cop, det, parataxis, punct)       \\
		Root(aux, cop, det, nsubj, parataxis, punct)   \\  \thickhline 
		\multicolumn{1}{c}{} \\\thickhline
		\multicolumn{1}{c}{Newsela}                  \\ \thickhline 
		Root(cop, nsubj, onl, punct)      \\
		Root(iobj, nsubj,punct,xcomp)    \\
		Root(advmod, cop, csubj, punct)      \\   \thickhline 
	\end{tabular}

	\caption{Examples of complex syntactic root rules.}
	\label{tbl.su}
	
\end{table}

We should point out that lexical and syntactic constraints can be
easily combined by merging the two sets of constraints provided by the
user. In this case six indicator features are used, three for the
lexical constrains and three for the syntactic constraints.

\section{Experimental Setup}
\label{sec:experimental-setup}

\paragraph{Datasets}
We experimented with two simplification datasets: (1)~Newsela
\cite{xu2015problems}, a simplification corpus of news
articles created by Newsela's professional editors. Each news
article is written at four different simplicity levels. It
consists of 1,130 articles, 30 of which are reserved as test
set; and (2)~WikiLarge \cite{zhang2017sentence}, a large
(296,402 sentence pairs) corpus which consists of a mixture of
three Wikipedia simplification datasets collated by
\citet{zhu2010monolingual}, \citet{woodsend2011learning} and
\citet{kauchak2013improving}. The test set for WikiLarge was
created by \citet{xu2016optimizing} and consists of~359
sentences, taken from Wikipedia, and then simplified using
Amazon Mechanical Turkers to create eight references per
source sentence.

\paragraph{Model Configuration}

For both datasets we used the Transformer as implemented within
OpenNMT-py \cite{opennmt}.  The encoder and decoder consist of 8
layers with a hidden dimension of size 500. Word embeddings, size 500,
were initialized randomly and shared between the encoder and decoder.
We used ten attentional heads and a copy  mechanism
\cite{see2017get}. The network was optimized using Adam
\cite{kingma2014adam} and SARI \cite{xu2016optimizing} was used for
early stopping.  The vocabulary size was limited to the~50,000 most
frequent tokens, the remaining tokens were replaced with an UNK token.

\begin{table}[t]
	\centering
	\begin{small}
		\begin{tabular}{@{~}l@{\hspace{1.4ex}}@{~}c@{\hspace{1.4ex}}@{~}c@{\hspace{1.4ex}}@{~}c@{\hspace{1.4ex}}}  \thickhline 
			Metric & SpencerK    & Newsela & WikiLarge \\ \thickhline
			Size  & 2,120  & 9,758    & 45,338     \\
			average(\#target) & 1.00     & 8.02    & 8.61      \\
			Recall    & 100\% & 23\%    & 21.1\%   \\ \thickhline
		\end{tabular}
	\end{small}
	\caption{Lexical substitution dictionaries, created
          manually (SpencerK is a shorthand for Spencer Kelly) or using GIZA++ on {Newsela} and
          {WikiLarge}.  Size: Number of source words in 
          dictionary. average(\#target): Average number of simplifications per
          source word. Recall: The recall of the dictionary calculated against
           SpencerK dictionary.  } 
	\label{tbl.dict}

\end{table}

\paragraph{Constraint Configuration}
Table~\ref{tbl.dict} presents statistics of the dictionaries used in
our experiments.  At test time, we explored two approaches to applying
the constraints to the encoder. For WikiLarge, simple tokens were
marked with keep and complex tokens were marked with replace. When
using the complex list, we included $\sim$12,000 most complex
words. For Newsela, simple tokens were marked with indifference and
complex tokens were marked with replace. When using the complex list,
we included $\sim$7,000 most complex words.  In both approaches, all
functions words were marked with indifference.

At test time, complex syntactic rules were marked with the replace
indicator and all other rules were always marked with the keep
indicator. For Newsela, when using the complex list, we include
approximately 29\% of the rules.  Whereas for WikiLarge we include
approximately 13\% of the rules.

\paragraph{Evaluation Metrics}

As there is no single agreed-upon metric for simplification,
we evaluated model output using the combination of five
automatically generated scores:\footnote{Our evaluation
	procedure can be found at \url{https://github.com/Jmallins/CROSS}.}

\begin{itemize}

\item \emph{BLEU} \cite{papineni-EtAl:2002:ACL}
assesses the degree to which generated simplifications
differ from gold standard references; unlike
\citet{zhang2017sentence}, we use
\texttt{multi-bleu.perl}\footnote{
	\citet{zhang2017sentence} use \texttt{mtevalv13a.pl} which is intended for untokenized
	text.}, as the test sets are already tokenized.

\item \emph{SARI} \cite{xu2016optimizing} is
calculated using the average of three rewrite operation
scores: addition, copying, and deletion. It rewards
addition operations when the system's output is not in
the input but occurs in the references; analogously, it
rewards words deleted/retained if they are in both the
system output and the references; our SARI
implementation differs from previous
versions\footnote{Further fixes were applied to cases
	where a single reference was provided.}, as we use the
precision of the delete operation when calculating SARI,
as recommended in \citet{xu2016optimizing}.  Previous
approaches used the F1 of all three rewrite operations.

\item  \emph{FKGL} the Flesch-Kincaid Grade Level index
measures the readability of the output (lower FKGL
implies simpler output). We modified FKGL such that a
newline indicates the end of sentence, so as to prevent
unrelated lines being calculated as one continuous
sentence.

\item \emph{S-BLEU} is a shorthand for self-BLEU and
computes the BLEU score between the output and the
source. This metric allows us to examine whether the
models are making trivial changes to the input.

\item \emph{Copy} measures the percentage of sentences
copied (with no changes made) from the source to the
output as a way of quantifying the extent to which a
model performs any rewriting at all.
\end{itemize}

We also evaluated system output by eliciting human
judgments via Amazon's Mechanical Turk.  Native English
speakers (self reported) were asked to rate
simplifications on three dimensions: Grammaticality (is
the output grammatical and fluent?), Meaning Adequacy
(to what extent is the meaning expressed in the original
sentence preserved in the output, with no additional
information added?), and Simplicity (is the output a
simpler version of the input?). The ratings were
obtained using a five point Likert scale.  100 sentences
were randomly sampled from the test set\footnote{We used
	the same samples as \citet{zhang2017sentence}.}, each
sample received five ratings, resulting in 500~judgments
per test set.

\begin{table}[t]
	\begin{center}	
		\begin{small}
			\begin{tabular}{@{~}l@{\hspace{1.44ex}}c@{\hspace{1.44ex}}c@{\hspace{1.44ex}}c@{\hspace{1.44ex}}c@{\hspace{1.44ex}}c@{\hspace{1.4ex}}} \thickhline 
				WikiLarge               & SARI & BLEU & FKGL & S-BLEU & Copy \\ \thickhline
				Reference & N/A & N/A & 8.24 & 63.92 & 16.2\% \\
				Source              &   26.31   &   {99.37}   & 9.54     & \hspace*{-.2cm}100.00       & 100\%     \\
				Truncate            & 35.62     &  99.32   &   9.54   &  95.48  &         0\% \\\hline 
				{PBMT-R}              &   40.30   &  81.02    &  8.40    &    74.95       &  09.7\%         \\
				Hybrid              &   27.59   &  48.69   &{4.72}    &    {30.57}       &  {03.1\%}         \\
				SBMT-SARI              &  40.75    & 73.01     &   7.53   &   67.93        & 10.6\%    \\\hline 
				EncDecA               &   39.58   & {89.00}     & 8.61     &  83.81         &    40.7\% \\
				DRESS               &  35.45   &  77.32&    6.76  &  56.96    &        21.5\%      \\  
				DRESS-Ls               &   36.08   & 80.35
				&  6.90    &   60.21        &   26.2\% \\ \hline
				Transformer         & 36.21     &   81.51   &   8.73   &      76.33     &  36.2\%         \\
				DMASS               & {40.35}     & 79.68     &  7.45    &    70.82       &  15.6\%         \\  
	
				CROSS-Lex &     38.82 &70.70      & 7.92      & 65.62          & 10.6\%           \\
			
				CROSS-Syn      & 33.89     &  64.98    &  7.98       &  68.88   & 19.9\%     \\
				CROSS               &  36.07    &  64.64     &7.46      & 56.11          & 15.6\%           \\ \thickhline 
				\multicolumn{6}{c}{} \\\thickhline
				Newsela               & SARI & BLEU & FKGL & S-BLEU & Copy \\ \thickhline
				Reference              & N/A    &  N/A     &3.43     & 17.81       & {0\%}     \\
				Source              &  11.97    &  20.79     &8.61      & \hspace*{-.2cm}100.00       & 100\%     \\
				Truncate            &   36.92   & 21.54     & 5.57     & 62.54           &  {0\%}         \\\hline 
				
				PBMT-R               & 41.23     & 17.62      & 7.96    &75.29         &  05.9\%         \\ 
				Hybrid               & 35.37     & 10.87     & 4.14   &19.96         &  03.3\%         \\\hline 
				EncDecA               & 42.98    & 21.17     & 5.48   &52.54         &  15.7\%         \\
				DRESS               & 42.85     & 22.65      & 4.20     & 39.69          &  11.3\%         \\ 
				DRESS-Ls               & {43.26}     & {23.66}      & 4.36     & 42.72          &  14.5\%         \\
				
				\hline 
				Transformer         &   42.21   & 19.90     &4.77      &   40.05        &  11.6\%         \\
				DMASS               & 37.36     &   07.51   &    3.84 & {11.15}          & {01.1\%}         \\

				CROSS-Lex &   41.56   &  18.88     &    3.81  & 33.98          &    06.8\%       \\

				CROSS-Syn &   38.12   &     14.30  & 3.48     & 21.35           & 05.1\%          \\
				CROSS               &  37.57    &  12.68     &{3.51}      & 26.55          & 05.6\%          \\ \hline 
			\end{tabular}
		\end{small}
	\end{center}

	\caption{Automatic evaluation on  WikiLarge and Newsela test
		set.  We also report the average FKGL, S-BLEU, and Copy of all
		references (Reference).}
	\label{tbl.AR}

\end{table}

\section{Results}

Our first suite of experiments compares our approach against
the state-of-the-art aiming to show that our model can also
function as a general-purpose simplification system. There is
no point having a controllable model if it cannot generate
adequate simplifications on its own.  Our second suite of
experiments examines how the simplicity level can be
manipulated.

\paragraph{Automatic Evaluation} 

Table~\ref{tbl.AR} summarizes our automatic evaluation
results\footnote{As our automatic metrics differ from previous papers
	we re-calculate all scores for all available simplification models.}
on WikiLarge and Newsela. We compared our model against three
well-established non-neural models: {PBMT-R}
\cite{wubben2012sentence}, a phrase-based machine translation model,
{SBMT-SARI} \cite{xu2016optimizing}, a syntax-based translation model
trained on PPDB and which is then tuned using SARI, and {Hybrid}
\cite{narayan2014hybrid}, a model which performs sentence spiting and
deletions and then simplifies with {PBMT-R}. We also compare against
various neural simplification models: (a)~the three LSTM-based models
reported in \citet{zhang2017sentence}, namely {EncDecA}, an
encoder-decoder model with attention, {DRESS}, a variant of {EncDecA}
trained with reinforcement learning, and its extension {DRESS-Ls}
which has an additional lexical simplification component; (b)~{DMASS}
\cite{zhao2018integrating}, a transformer-based model enhanced with
simplification rules from PPDB; and (c)~a vanilla transformer-based
encoder-decoder model without any constraints.

We report results for several variants of our model which we call
{CROSS} as a shorthand for \textbf{C}ont\textbf{RO}llable
\textbf{S}entence \textbf{S}implification. \mbox{{CROSS-Lex}} contains
lexical constraints only, \mbox{{CROSS-Syn}} focuses solely on
syntactic simplifications, while {CROSS} is the full model with both
types of constraints. For the sake of brevity, we only report results
with constraints provided in a list format as these performed slightly
better on the development set. We also include two strong baselines,
repeating the source sentence (Source) and truncating the source
sentence to the first $N$~words, as determined by the validation set
(Truncate).

\begin{table}[t]
	\begin{center}
		\begin{small}
			\begin{tabular}{@{~}l@{\hspace{1.2ex}}l@{\hspace{1.2ex}}l@{\hspace{1.2ex}}l@{\hspace{1.2ex}}l@{\hspace{1.2ex}}l@{\hspace{1.2ex}}}\thickhline
				WikiLarge     & Gram & Mean & Simp & AVG & Min \\ \thickhline 
				Reference     &    4.01*    &    4.13**     & 3.56**           &  3.90**   &  3.16*   \\
				
				DRESS-Ls     &    4.32**     &3.97**         & 3.14           &3.81**     & 2.80    \\
				DMASS     &   3.69      &  3.21       & 2.57**           &    3.16 &2.29**     \\ 
				Transformer     &    3.91     &  3.63       & 3.04**           &    3.53 &  2.72**   \\
		
				CROSS-Lex   &      3.72   & 3.41        & 3.18           &3.43     &  2.80   \\
				CROSS-Syn   &   3.54      & 2.22        & 2.46**            &3.07*     &2.15**     \\
				CROSS &    3.61    & 3.37        &   3.13
				&  3.37   &    2.84\\ 
				
				\thickhline
				
				\multicolumn{6}{c}{} \\ \thickhline
				Newsela     & Gram & Mean & Simp & AVG & Min \\ \thickhline 
				Reference     &  4.11**       & 3.73**     & 3.88**       &3.91**    &  3.47**    \\
				
				DRESS-Ls     &     3.33*  & 2.98**     &2.93          &3.08**  & 2.45   \\
				DMASS     &  2.05**  & 1.55** & 1.74**          &1.78**     & 1.39**     \\ 
				Transformer     &  2.88**     & 2.47**      & 2.70       &2.68**     & 2.00**     \\

				CROSS-Lex   &    3.07**    &2.89**        & 2.95           &2.97**    & 2.45   \\
				CROSS-Syn   &     3.60    &      3.37   &2.89            &3.27     & 2.31    \\
				CROSS &    3.54     &    3.41     &   2.91
				& 3.28    & 2.29    \\ 
				
				\thickhline
			\end{tabular}
		\end{small}
	\end{center}

	\caption{\label{tab:humans}Human evaluation on WikiLarge and Newsela. Models
		significantly different from CROSS are marked with * $(p <
		0.05)$ and ** $(p < 0.01)$. Significance tests were
		performed using a student \emph{t}-test.}

\end{table}

Results on WikiLarge are mixed, with no model being best for every
metric. We see that SBMT-SARI achieves the highest SARI, with minimal
copying and a moderate \mbox{S-BLEU}. Of the two existing
state-of-the-art models, DRESS-Ls and DMASS, we see that DRESS-Ls
achieves moderate SARI and S-BLEU scores, however, it has a high Copy
score. This suggests that DRESS-Ls is very polar, applying high
amounts of rewriting to some sentences and keeping others completely
unchanged.  DMASS achieves the second highest SARI score and Copy is
low, however, \mbox{S-BLEU} is high suggesting it produces modest changes
consistently.

CROSS achieves a slightly worse SARI than the baseline
Transformer, however this is in part due to the Transformer's
high Copy and high S-BLEU. In contrast, CROSS achieves a low
\mbox{S-BLEU} and Copy score similar to that of the
references. CROSS has a lower SARI compared to CROSS-Lex,
however, it has a better S-BLEU and Copy. CROSS outperforms
CROSS-Syn with a better SARI and Copy score.  The results also
show that \emph{standard} encoder-decoder models (EncDecA,
Transformer) produce outputs which are highly similar to the
input, highlighting the importance of constraining the output.

We next consider the Newsela dataset.  We see that DRESS-Ls
achieves the highest SARI, however, it also has the highest
level of copying and a moderately high S-BLEU. DMASS, on the
other hand, achieves a low SARI, but with a low amount of
copying and a low S-BLEU. Also notice that the Truncate
baseline has the highest BLEU score, outside of the DRESS
models.  The Transformer achieves a moderate SARI, however, it
also has a high Copy and high S-BLEU.  CROSS achieves a low
SARI which in part can be explained by its high level of
rewriting as seen in the low S-BLEU and Copy.  We see that
CROSS-Lex has a higher SARI compared to CROSS but worse S-BLEU
and Copy scores.  CROSS-Syn and CROSS both have very similar
scores, however, CROSS-Syn performs more
rewrites.

\paragraph{Human Evaluation} 
The results of our human evaluation are presented in
Table~\ref{tab:humans}. We follow previous approaches and
report Grammaticality, Meaning Adequacy, and Simplicity
individually and combined (AVG is the average of the three
dimensions). In addition, we include a new metric
\emph{Minimum}, which is the (average) minimum value of
Grammaticality, Meaning Adequacy, and Simplicity per sentence.
We include Minimum because we argue that a simplification is
only as good as its weakest dimension. We note that it is
trivial to produce a sentence that is perfectly adequate and
fluent, by simply repeating the source sentence. It is also
easy to produce a simple sentence if we do not care about
adequacy. We evaluated CROSS (and CROSS-Lex, CROSS-Syn
variants) against the two state-of-the-art models {DMASS} and
{DRESS-Ls} as well a Transformer baseline. We also elicited
judgments on the gold standard Reference as an upper bound.

Human evaluation on WikiLarge (top half in
Table~\ref{tab:humans}) shows that both DRESS-Ls and CROSS
achieve highest scores for Minimum. CROSS significantly
outperforms all other models for both Min and Simplicity.
Transformer achieves a higher score for both Grammaticality
and Meaning compared to CROSS. However, this can be explained
due to the high Copy score, which therefore guarantees high
Grammaticality and Adequacy scores.  This can also in part
explain the high Grammaticality and Meaning Adequacy scores
for \mbox{DRESS-Ls}. CROSS-Syn achieves lower scores compared
to CROSS-Lex, suggesting that syntactic changes are not as
important for WikiLarge.

Human evaluation on Newsela (second half of Table~\ref{tab:humans})
shows that all CROSS variants are better than related Transformer and
DMASS models across all metrics. CROSS and DRESS-Ls both achieve the
highest Minimum scores. For all other metrics, CROSS is better or the
same than all other models.  CROSS and CROSS-Syn achieve similar
results, both outperforming CROSS-Lex.  This suggests that syntactic
simplifications are more prominent in Newsela compared to WikiLarge.

\begin{table}[t]
	\begin{small}
		\begin{center}
			\begin{tabular}{@{~}l@{\hspace{1.2ex}}c@{\hspace{1.2ex}}c@{\hspace{1.2ex}}c@{\hspace{1.2ex}}c@{\hspace{1.2ex}}c@{\hspace{1.2ex}}c@{\hspace{1.2ex}}}\thickhline
				& Lex & Syn & Voice & Tense & Split & All \\ \thickhline 
				Reference & 35\% & 10\% & 7\% & 5\% & 6\% & 41\%\\  
				Transformer & \hspace*{.8ex}9\%&\hspace*{.8ex}1\% &\hspace*{.2ex}1\% &0\% &0\% &\hspace*{.8ex}9\% \\
				CROSS &  29\%  & \hspace*{.8ex}8\% & 8\%    & 2\% & 8\% & 35\%\\ \thickhline
	
			\end{tabular}
		\end{center}
	
		\caption{\label{tab:analysis} Proportion of simplifications on a
			100 sentence sample from the WikiLarge and Newsela test sets.}
	\end{small}
	
\end{table}

\begin{table}[t]
	\begin{center}
		\begin{small}
			\begin{tabular}{@{~}l@{\hspace{1.2ex}}l@{\hspace{1.2ex}}l@{\hspace{1.2ex}}l@{\hspace{1.2ex}}l@{\hspace{1.2ex}}l@{\hspace{1.2ex}}l@{\hspace{1.2ex}}}\thickhline
				WikiLarge     & Gram & Mean & Simp & AVG & Min &FKGL\\ \thickhline 
				XSimple   &  3.30      &    3.09    & 3.06*           &  3.15  & 2.84  & 6.96 \\
				Simple &       3.24  &3.11         & 2.87           &     3.08 & 2.77 & 7.46     \\ \thickhline
				\multicolumn{7}{c}{} \\ \thickhline
				Newsela     & Gram & Mean & Simp & AVG & Min
				& FKGL\\ \hline 
				XSimple   &  3.46**     &    2.88**    & 3.11**           &  3.15  & 2.33**&  2.91   \\
				Simple &     3.89  & 3.59         & 2.53           &     3.34 & 2.10   &  3.51 \\ \thickhline
			\end{tabular}
		\end{small}
	\end{center}

	\caption{Human evaluation on varying simplicity of
		model output. Ratings that are significantly
		different are marked with * $(p < 0.05)$ and ** $(p <
		0.01)$. Significance tests were performed using a student
		\emph{t}-test.} 
	\label{tbl.var}

\end{table}

\begin{table}[t]
	\centering
	\begin{footnotesize}	
		\begin{tabular}{@{}l@{~~}p{7cm}@{~} }
			\thickhline
			{Complex}& In its pure form, Dextromethorphan occurs as a white powder.\\ 	
			{Reference} & Dextromethorphan {is} a white powder in its pure form. \\ 
DRESS-Ls & In its pure form, Dextromethorphan occurs as a white powder.\\
			{Simple} &  In its pure form, Dextromethorphan {is like} a white \mbox{powder.}\\
			{XSimple} &Dextromethorphan {can be found} as white powder. \\
			\hline 
	
			{Complex} & The Pentagon is poised to spend billions to build a new stealth bomber, a top secret project that could bring hundreds of jobs to the wind-swept desert communities in Los Angeles County's northern reaches.\\ 	
			{Reference} & Mission to build the secret
                        warplane. \\
DRESS-Ls &  The Pentagon is trying to spend billions to build a new drone.\\
			{Simple} &  The Pentagon  secret project that could bring hundreds of jobs to the desert-swept communities in Los Angeles County. \\
			{XSimple} &  It could also bring hundreds of jobs. \\
			\hline 
			{Complex} &  The United States is about to spend billions of dollars to build a top-secret warplane.\\ 	
			{Reference} & Mission to build the secret warplane \\ 
DRESS-Ls & The United States is about to spend billions of dollars to build a secret bomb.\\
			{Simple}&    The United States is about spend dollars to build a top-secret warplane.\\
			{XSimple} & The United States is about to build a warplane. \\
			\thickhline 
			
		\end{tabular}
	\end{footnotesize}

	\caption{Output of Simple and XSimple systems based on
          CROSS. We also show the output of DRESS-Ls, the source
          complex sentence, and the
          reference simplification.}
	\label{tbl:example}

\end{table}

\paragraph{Analysis of Model Output} We further analyzed the
simplifications produced by CROSS to gain insight on the types
of simplifications it generates. We sampled 100 sentences (50
from each test set) and classified the simplifications into
two categories, namely lexical (Lex) or syntactic (Syn). For
syntactic simplifications we further marked whether these
pertained to common changes, i.e.,~passive to active voice
(Voice), past tense to present or past perfect (Tense), and
sentence splitting (Split). Table~\ref{tab:analysis} shows a
breakdown of these phenomena for CROSS, the baseline
Transformer model, and the references. As can be seen, CROSS
performs similar simplifications to the references, and
substantially more syntactic changes compared to the Transformer.

\paragraph{Controllability} A central claim of this paper is
that CROSS can be adapted to user needs.  We test this claim,
by experimenting with varying the simplicity level of the
output. Specifically, we sampled 100 complex source sentences
(with FKGL score of~11 or higher) from the WikiLarge and
Newsela test sets and produced two sets of outputs, one with
our general-purpose system which produces a moderate amount of
simplification (Simple), and another one where we forced the
model to simplify more drastically, extra simple
(XSimple). This was achieved by increasing the number of
lexical and syntactic constraints the model must adhere
to. Specifically, we include the~12,000 most complex words for
Newsela, and the~18,000 most complex tokens for Wikilarge. We
also increased the number of complex syntactic constraints to
approximately 40\%~for Newsela and 25\%~for WikiLarge.

Results in Table~\ref{tbl.var} show that CROSS is able to successfully
alter the simplicity level of the output. For both datasets we see
that participants perceive differences between the output of the
simple and XSimple models (this is also reflected in the FKGL which is
lower for XSimple).  For WikiLarge, all scores apart from
simplification do not differ significantly. For Newsela, we see that
XSimple sentences are significantly less adequate and
grammatical. However, on average Simple and XSimple sentences do not
significantly differ, showing a trade-off between simplicity and
adequacy/grammaticality. Examples of system output are shown in
Table~\ref{tbl:example} (and in the supplementary).

\section{Conclusions}

We developed a simplification model, which is able to jointly or
individually control the syntax and lexical choice of its
output. Experiments showed that our constraint-aware model produces
state-of-the-art simplification results, receiving the best Minimum
score on WikiLarge. We further showed that by adjusting these
constraints we are able to control the level of simplification of the
output. In the future we plan to incorporate more explicit controls,
e.g., allowing the user to determine if the sentence should be split
or not, or the readability level of the output.  \small
\bibliography{emnlp-ijcnlp-2019}

\clearpage
\section{Appendix}

Table~\ref{tbl.cw} presents examples of the lists of complex words and
dictionaries used in our experiments. We also show simplification
examples created by our model and comparison systems on WikiLarge and
Newsela (see Tables~\ref{tbl:example1}--\ref{tbl:example:final}). 

\begin{table}[h]
	
	\centering
	\begin{tabular}{ll|ll} 
		\multicolumn{2}{c}{List} &   \multicolumn{2}{c}{Dictionary} \\ \thickhline 
		WikiLarge    & Newsela   &  Complex & Simple \\ \thickhline 
		kunaich      & expans    & abandon        & leave         \\
		ufficial     & personnel & assembled      & built         \\
		insofar      & epidem    & cherishing     & loving        \\
		silkmoth     & array     & customary      & normal        \\
		lagerphonist & deploy    & educating      & teaching      \\
		conduc       & swath     & fraudulent     & fake          \\
		offbeat      & perspect  & initiated      & started       \\
		cavalier     & dispar    & iterated       & repeated      \\
		midfield     & crucial   & replenished    & refilled      \\
		straddl      & prototyp  & shove          & push         \\ \thickhline  
	\end{tabular}
	
	\caption{Examples of (stemmed) complex words and complex-simple dictionary. }
	\label{tbl.cw}
\end{table}
\vspace*{-.5cm}
\begin{table*}[h!]
	\centering
	
	\begin{tabular}{| l  p{13cm}@{~} |}
		\hline
		Complex & It is \st{situated} at the coast of the Baltic Sea, where it encloses the city of Stralsund. \\ 	
		Reference & It is \textbf{located} at the coast of the Baltic Sea where it \textbf{surrounds} the city of Stralsund. \\ 
		DRESS-Ls & It is \textbf{situated} at the coast of the Baltic Sea.  \\  
		Transformer & It is \textbf{situated} at the coast of the Baltic Sea, where it encloses the city of Stralsund. \\ 
		DMASS & It is \textbf{located} at the shore of the Baltic Sea, where it is located at the shore of the borough of Stralsund. \\
		{CROSS-Lex} &  It is \textbf{at} the coast of the Baltic Sea, where it encloses the city of Stralsund. \\
		{CROSS-Syn} & It is \textbf{located} at the coast of the Baltic Sea. \\
		{CROSS} &  It is \textbf{found} at the coast of the Baltic Sea, near the city of Stralsund. \\ \hline 
		\hline
		Complex & In 1987 Wexler was \st{inducted} into the Rock and Roll Hall of Fame.\\ 	
		Reference & In 1987 Wexler was inducted into the Rock and Roll Hall of Fame. \\ 
		DRESS-Ls & In 1987 Wexler was inducted into the  Rock and Roll Hall of Fame.  \\  
		Transformer & In 1987, Wexler was inducted into the  Rock and Roll Hall of Fame. \\ 
		DMASS & In 1987 Wexler was inducted into the  Rock and Roll Hall of Fame.\\
		{CROSS-Lex} &  In 1987 Wexler was \textbf{added into} the  Rock and Roll Hall of Fame. \\
		{CROSS-Syn} &  \textbf{He} was inducted into the Rock and Roll Hall of Fame in 1987. \\
		{CROSS} &  \textbf{He} was \textbf{added into} the  Rock and Roll Hall of Fame in 1987. \\
		\hline
		
	\end{tabular}
	\caption{System output on WikiLarge.   We show the source
		Complex sentence, the Reference, and 
		output from DRESS-Ls, a Transformer, DMASS, and three CROSS
		variants; the full system (CROSS), and with on lexical 
		(CROSS-Lex) and syntactic CROSS-Syn) constraints. Items signalled for replacement
		are marked with a \st{strike out} and substitutions are 
		in \textbf{bold}.}
	\label{tbl:example1}
\end{table*}
\vspace*{-.2cm}

\begin{table*}[h!]
	\centering
	
	\begin{tabular}{| l  p{13cm}@{~} |}
		\hline
		Complex & In its pure form, Dextromethorphan occurs as a white powder.\\ 	
		Reference & Dextromethorphan \textbf{is} a white powder in its pure form. \\ 
		DRESS-Ls & In its pure form, Dextromethorphan occurs as a white powder. \\ 
		{Simple} &  In its pure form, Dextromethorphan \textbf{is like} a white powder.\\
		{XSimple} & Dextromethorphan \textbf{can be found} as white powder. \\
		\hline \hline
		Complex & After graduation he returned to Yerevan to teach at the local Conservatory and later he was {appointed} artistic director of the Armenian Philarmonic Orchestra.\\ 	
		Reference & After graduation he returned to Yerevan to teach at the local Conservatory and was later appointed artistic director of the Armenian Philarmonic Orchestra. \\ 
		DRESS-Ls &  After graduation he returned to Yerevan to teach at the local Conservatory.  \\
		{Simple} &  After graduation he returned to Yerevan to teach at the local Conservatory and later he \textbf{became} artistic director of the Armenian Philarmonic Orchestra. \\
		{XSimple} &  After graduation he returned to Yerevan to teach at the Conservatory and later he \textbf{became} artistic director of the Armenian Philarmonic Orchestra. \\
		\hline 
		
	\end{tabular}
	\caption{System output on WikiLarge for varying  simplicity levels. We show the source
		Complex sentence and the Reference as well as
		output from DRESS-Ls, and two variants of our model Simple
		and XSimple. Substitutions are shown in \textbf{bold}. }
	\label{tbl:example2}
\end{table*}

\begin{table*}[h]
	\centering
	
	\begin{tabular}{| l  p{13cm}@{~} |} \hline 
		
		Complex & He thinks the new \st{stealth} bomber program would \st{ultimately} cost \$90 billion.\\ 	
		Reference & He thinks the new stealth bomber program would \textbf{actually} cost \$90 billion.\\ 
		DRESS-Ls & He thinks the \textbf{new combat number} would cost \$90 billion.  \\  
		Transformer & He thinks the new bomber program would cost \$90 billion.\\ 
		DMASS & The new \$ sinkhole program would be used this year.\\
		{CROSS-Lex} &  He thinks the new program would cost \$90 billion.\\
		{CROSS-Syn} &  He thinks the program would cost \$90 billion for the new stealth bomber. \\
		{CROSS} & He thinks the program would cost \$90 billion for the new bomber. \\
		\hline \hline
		Complex & But then he heard the radio : a \st{massive} glacier had crashed down the mountain.
		\\ 	
		Reference & A huge glacier of ice had just crashed down the mountain.\\ 
		DRESS-Ls &  But then he heard the radio: A massive glacier had crashed down the mountain.
		\\  
		Transformer & But then he heard the radio: A massive glacier had crashed down the mountain.\\ 
		DMASS & Then he heard the radio even though a \textbf{huge piece} had crashed down the mountain.\\
		{CROSS-Lex} &  But then he heard the radio: A \textbf{huge} glacier had crashed down the mountain.\\
		{CROSS-Syn} &  But then he heard a glacier crash down the mountain. \\
		{CROSS} & But then he heard a glacier crash down the mountain. \\
		\hline
		
	\end{tabular}
	\caption{System output on Newsela. We show the source Complex sentence and the  Reference as well as output
		from DRESS-Ls, a  Transformer, DMASS, and three variants of our model; CROSS is the full system,
		CROSS-Lex applies only lexical constraints, while CROSS-Syn only
		syntactic ones. Substitutions are shown in \textbf{bold}. Lexical items indicated for replacement are marked with a \st{strike out}.}
	\label{tbl:example3}
	
\end{table*}


\begin{table*}[h]
	
	\centering
	
	\begin{tabular}{| l  p{12cm}@{~} |}
		\hline
		Complex & The Pentagon is poised to spend billions to build a new stealth bomber, a top secret project that could bring hundreds of jobs to the wind-swept desert communities in Los Angeles County's northern reaches.\\ 	
		Reference &Mission to build the secret warplane. \\ 
		DRESS-Ls &  The Pentagon is trying to spend billions to build a new drone.  \\
		{Simple} & The Pentagon  secret project that could bring hundreds of jobs to the desert-swept communities in Los Angeles County. \\
		{XSimple} & It could also bring hundreds of jobs. \\
		\hline \hline
		Complex & The United States is about to spend billions of dollars to build a top-secret warplane.\\ 	
		Reference & Mission to build the secret warplane. \\ 
		DRESS-Ls & The United States is about to spend billions of dollars to build a secret bomb. \\ 
		{Simple} &   The United States is about spend dollars to build a top-secret warplane.\\
		{XSimple} & The United States is about to build a warplane. \\
		\hline

	\end{tabular}
	\caption{System output on Newsela for varying  simplicity levels. We show the source Complex sentence and
		the  Reference as well as output from DRESS-Ls, and two
		variants of our model Simple and XSimple. Substitutions are shown in \textbf{bold}. }
	\label{tbl:example:final}
\end{table*}

\end{document}